\documentclass[conference]{IEEEtran}
\IEEEoverridecommandlockouts
\usepackage{cite}
\usepackage{amsmath,amssymb,amsfonts}
\usepackage{algorithmic}
\usepackage{graphicx}
\usepackage{textcomp}
\usepackage{xcolor}

\def\BibTeX{{\rm B\kern-.05em{\sc i\kern-.025em b}\kern-.08em
    T\kern-.1667em\lower.7ex\hbox{E}\kern-.125emX}}
\begin{document}

\title{Counterfactual Explanations of Black-box Machine Learning Models using Causal Discovery with Applications to Credit Rating\\
{\footnotesize \textsuperscript{*}}
\thanks{}
}
\author{\IEEEauthorblockN{Daisuke Takahashi}
\IEEEauthorblockA{\textit{Graduate School of Data Science,} \\
\textit{Shiga University, Hikone, Japan} \\
daisuke199968@gmail.com}
\and
\IEEEauthorblockN{Shohei Shimizu}
\IEEEauthorblockA{\textit{Graduate School of Data Science,} \\
\textit{Shiga University, Hikone, Japan}\\
\textit{RIKEN Center for Advanced }\\
\textit{Intelligence Project, Tokyo, Japan} \\
shohei-shimizu@biwako.shiga-u.ac.jp}
\and
\IEEEauthorblockN{Takuma Tanaka}
\IEEEauthorblockA{\textit{Graduate School of Data Science,} \\
\textit{Shiga University, Hikone, Japan} \\
takuma-tanaka@biwako.shiga-u.ac.jp}
}

\maketitle

\begin{abstract}
Explainable artificial intelligence (XAI) has helped elucidate the internal mechanisms of machine learning algorithms, bolstering their reliability by demonstrating the basis of their predictions. Several XAI models consider causal relationships to explain models by examining the input-output relationships of prediction models and the dependencies between features. The majority of these models have been based their explanations on counterfactual probabilities, assuming that the causal graph is known. However, this assumption complicates the application of such models to real data, given that the causal relationships between features are unknown in most cases. Thus, this study proposed a novel XAI framework that relaxed the constraint
that the causal graph is known. This framework leveraged counterfactual probabilities and additional prior information on causal structure, facilitating the integration of a causal graph estimated through causal discovery methods and a black-box classification model. Furthermore, explanatory scores were estimated based on counterfactual probabilities. Numerical experiments conducted employing artificial data confirmed the possibility of estimating the explanatory score more accurately than in the absence of a causal graph. Finally, as an application to real data, we constructed a classification model of credit ratings assigned by Shiga Bank, Shiga prefecture, Japan. We demonstrated the effectiveness of the proposed method in cases where the causal graph is unknown.

\end{abstract}


\section{Introduction}\label{S1}
In recent years, the use of artificial intelligence (AI) has rapidly increased owing to its revolutionary impact on various aspects of society and industry. This is primarily attributed to advances in deep learning \cite{deeplearn} and ensemble learning algorithms such as LightGBM \cite{ensmble}. However, it is impossible for humans to understand the basis for the predictions by these models due to the complex calculations and internal structures. Consequently, people are hesitant to utilize machine learning for tasks that require accountability (bank loans, medical diagnosis) \cite{AI-kadai}. To address this challenge, numerous 
methods have been developed in explainable artificial intelligence (XAI) to improve the explainability of black-box models \cite{XAI}.

Among the most promising approaches of XAI is the explanation of predictive models using causal inference \cite{causalXAI}. Causal inference-based XAI considers the correlation between the input and output and the dependence of features; thus, it can efficiently explain the desired predicted value. In particular, the estimation by existing methods, such as decision tree-based feature importance \cite{tree} and SHAP \cite{shap}, is based on the correlation of variables, which may be pseudo-correlation \cite{shapdame}. Most XAI methods based on causal inference employ counterfactual explanations, which elucidate a prediction by computing the change in the individual’s features to modify the prediction to the desired class \cite{CML}. A previous study \cite{LEWIS} presented an XAI system called LEWIS, which leveraged causal models and counterfactual probabilities. This study employed loan approval as an example to estimate a counterfactual probability score for a binary classification. Based on this explanatory score, this method rendered the features requiring change as explicit to modify the  prediction if the loan for a customer was predicted to be rejected.
 
 However, LEWIS requires complete knowledge of the causal graph of features. When building machine learning models, background knowledge such as causal relationships are rarely completely known. This is a major challenge when applying this method to real data. When the causal graph is unknown, it can be estimated by causal discovery \cite{CD}; however, its performance in artificial and real-world data has not been explored.
 
 \begin{figure*}[t]
    \centering
    \includegraphics[scale=0.55]{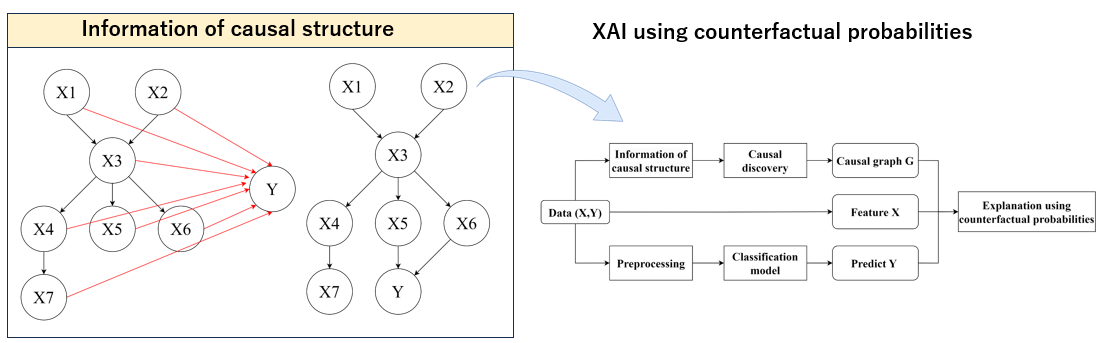}
    \caption{Framework of counterfactual probability explanations using causal structure information}
    \label{fig11}
\end{figure*}
 This study proposed a new XAI framework that relaxed the constraint of the causal graph being known. As shown in Figure~\ref{fig11}, a causal discovery can be possibly performed with prior knowledge on the causal structure, and the resulting causal graph can be used to explain the features using counterfactual probabilities. Our contributions are summarized as follows:
 \begin{itemize}
     \item We conducted numerical experiments to analyze the variation of explanatory scores with causal structure and the proposal of useful prior information on causal structure.
     \item Our artificial data experiments implied that the combination of causal discovery with certain prior information proposed above recovered the estimates of the explanatory score better than previous methods without causal discovery or graph assumption.
     \item By applying the method to real world financial data from the Shiga bank, Ltd., which is the largest regional bank in Shiga prefecture of Japan, we demonstrated that useful explanations could be made from the graph estimated by causal discovery. To the best of our knowledge, this is the first real-world example of counterfactual probability explanations in case that the causal structure is unknown and is estimated by causal discovery.
 \end{itemize}

 The remainder of this paper is organized as follows. Section II provides an overview of related research and defines the symbols used in this paper. Section III analyzes the effects of causal structures on explanation scores and proposes the use of useful prior information on the causal structure. In Section IV, using artificial data, we provide counterfactual explanations by combining prior information on 
 the causal structure and causal exploration. Consequently, we examine whether the estimated explanatory score can restore the estimate of the true explanatory score. Section V demonstrates that the proposed method is useful even when the causal graph is unknown using real data. Finally, Section VI summarizes the results and provides future perspectives.

\section{Background}\label{S2}
This section introduces the mathematical symbols and formulas used in this paper and outlines the related methods.
\subsection{Structural Causal Model}
Structural causal model (SCM) \cite{pearl2016} is a mathematical framework for handling causal relationships. SCM comprises a set of endogenous variables $\mathbf{V}=\{V_1,V_2,...,V_p\}$, a set of exogenous variables $\mathbf{U}=\{U_1,U_2,...,U_q\}$, which are variables not determined by the endogenous variables, and a set of functions $\mathbf{F}=\{f_1,f_2,...,f_p\}$, which determines the values of endogenous variables from those of other endogenous and exogenous variables. In addition, we define the parent set of endogenous variables as $\mathrm{pa}(V_i)$ and we assume variables in $\mathbf{U}$ are independent. Consequently, if SCM is specified by
\begin{equation}
    V_i = f_i(\mathrm{pa}(V_i),U_i)
    \label{scm}
\end{equation}
and the system is assumed to be autonomous, Equation~(\ref{scm}) is called a structural causal model, wherein the quantitative causal relations of variables are expressed by a directed graph called a causal graph. Here, autonomy implies that changing any function or distribution of variables does not change the distribution of other functions or distributions.
\subsection{LEWIS}
LEWIS \cite{LEWIS} is an XAI method that provides counterfactual explanations for machine learning predictions based on structural causal models. Consider the following binary classification problem. The explanatory variable is $X\in \mathbf{V}$ and the value of the explanatory variable is $\{x,x'\}\in X$ $(x>x')$, wherein all the values of the explanatory variables are discretized. Let $O\in\{o, o'\}$ be the predicted value corresponding to the explanatory variable, where $o$ is the positive predicted value and $o'$ is the negative predicted value. The counterfactual in any causal model $M=\langle V,U,F\rangle$ can be defined for a certain individual $u\in U$ as \cite{pearl2016}
\begin{equation}
 \label{cf}
    X(u)=x \Rightarrow  Y_{X=x}(u)=Y_{M_x}(u),
\end{equation}
where $Y_{X=x}$ is the counterfactual of the value of $Y$ if the value of $X$ were $x$, with $Y_{M_x}$ being that of $Y$ if the value of $X$ were $x$ in the causal model $M$. Further, the counterfactual probability $P(Y_{X=x})$ can be expressed as $P(Y|\mathit{do}(X=x))$ using the intervention symbol $\it{do}$. It represents the conditional probability of $Y$ if the value of $X$ were changed to $x$ and can be computed based on a causal graph.

In this setting, LEWIS defines the following three explanatory scores as the probability of a counterfactual,
necessity score (Nec)
\begin{equation}
      \label{nec} \mathrm{Nec}(x,x')=P(o'_{X=x'}|x,o),
  \end{equation}
sufficiency score (Suf)
\begin{equation}
 \label{suf}
 \mathrm{Suf}(x,x')=P(o_{X=x}|x',o'),
\end{equation}
and necessity and sufficiency score (Nesuf)
\begin{equation}
 \label{nesuf}
 \mathrm{Nesuf}(x,x')=P(o'_{X=x'},o_{X=x}).
\end{equation}
The necessity score and sufficiency score (called reversal probability) calculate the probability of the degree of change in the predicted value if the explanatory variable had a different value, given the value and predicted value of a given explanatory variable. In particular, the necessity score represents the probability that the predicted value would change if the value of the explanatory variable decreased, whereas the sufficiency score represents the probability that the predicted value would change if the value increased. The necessity and sufficiency score is a score that balances the necessity score and sufficiency score and can be considered as a feature importance in machine learning \cite{LEWIS}.

In \cite{LEWIS}, LEWIS was used to examine a binary classification model that determines whether a loan was approved or rejected. For example, $\text{Nec}=0.1$ for a feature implies that if a person whose loan is predicted to be approved decreased the value of a feature, there would be at most a 10\% chance of a prediction for the loan to be rejected. In addition, $\text{Suf}=0.8$ implies that if a person whose loan is predicted to be rejected increased the value of a certain feature, there is a maximum probability of 80\% that the loan would be predicted to be approved.

 These three scores are expressed as 
 \begin{gather}
      \label{nec2} \mathrm{Nec}(x,x')=\frac{P(o'|\mathit{do}(X=x'))-P(o'|x)}{P(o|x)},\\
      \label{suf2} \mathrm{Suf}(x,x')=\frac{P(o|\mathit{do}(X=x))-P(o|x')}{P(o'|x')},\\
 \label{nesuf2}
 \mathrm{Nesuf}(x,x')=P(o'|\mathit{do}(X=x'))-P(o'|\mathit{do}(X=x)),
\end{gather}
if the causal graph $G$ corresponding to the causal model $M$ is known and the monotonicity $(x>x'\Rightarrow O_x > O_{x'})$ is satisfied. In this case, the global explanation score of LEWIS, $\mathrm{maxNesuf}(X)$, is given by the maximum value of $\mathrm{Nesuf}(x,x’)$ for all pairs of the value of the explanatory variable, $(x,x’)$. Finally, if no causal graph is provided to LEWIS, it assumes that $P(O|\mathit{do}(X))=P(O|X)$ under the assumption that there are no confounding factors, which is likely to be violated.
 
\section{Numerical experiment of exploring useful prior knowledge on causal structure}\label{S3}
In this section, we analyze the effects of causal structures on explanation scores and propose useful prior information on the causal structure.
\subsection{Analysis of the influence of causal structure on explanation scores}\label{a1}
\subsubsection{Experimental design}
\begin{figure}[t]
    \centering
    \includegraphics[scale=0.6]{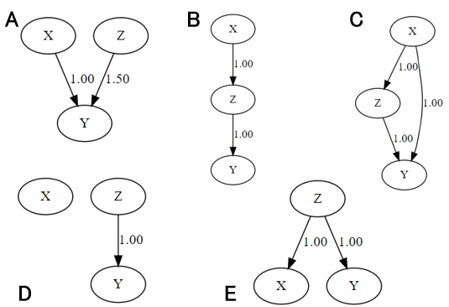}
    \caption{Causal graph used in analysis. The values on the directed edges represent the coefficients of the respective structural equations.}
    \label{fig:ABCDE}
\end{figure}

We built a machine learning model leveraging data generated from a three-variable causal structure and analyzed the characteristics of the LEWIS explanation score estimated for the predicted value. Data were generated from the five causal graphs in Figure~\ref{fig:ABCDE}. Structures A, B, and E are termed the collider, chain, and fork paths, respectively, and each exhibited its own unique independence \cite{pearl2016}. Structure C exhibited confounding behavior, where the variable $X$ was the confounding variable. Structure D contained a variable $X$ that was independent of variable $Y$. The coefficients of each structural equation were all equal for each variable in structures B–E, and the coefficients of A were set to 1 and 1.5 for $X$ and $Z$, respectively. After generating data from the five causal structures and building a machine learning model using that data, we estimated the Nesuf score of Equation~(\ref{nesuf}) with LEWIS. This was iterated 100 times and the characteristics of the average estimated value were analyzed. 

\begin{table}[ht]
    \centering
    \caption{Experimental conditions}
    \label{tab:exp1}
    {
    \begin{tabular}
    {|cc|}
    \hline
         Function&  \{linear, nonlinear\}\\
         Probability of error variables & uniform distribution of $[0,1]$\\
         Model& catboost\\ 
         Sample size& 5000\\
         Discretization method & equal-width discretization\\
         Number of discretization bins & 10\\

         \hline
    \end{tabular}
    }
\end{table}\par
Table~\ref{tab:exp1} presents the detailed experimental settings. In each repetition, error variables were generated from a uniform distribution of $[0,1]$ using a linear or nonlinear (second-order monomial) function system for the causal structures A–E. The generated data were discretized using the same discretization bin number, and catboost was used as the classifier. Herein, the target variable $Y$ was also discretized to binary values using equally spaced discretization. The sample size of the data was $5\,000$, which Nesuf was estimated for the predictions of.

\subsubsection{Experimental results}
The estimates of Nesuf are presented in Tables~\ref{tab:nesufL} and~\ref{tab:nesufN}. The results suggested two possible pieces of prior information on causal structure to determine Nesuf based on a linear causal structure. First, in case of multiple variables that are direct parents of a target variable such as structure A, variables with strong correlations would be more important. In structure A, the correlation coefficients with $Y$ of variables $Z$ and $X$ were 0.8075 and 0.4779 on average, respectively, indicating that variable $Z$ exhibited a stronger correlation than $X$ and was therefore more important. Second, if the target variable is independent and has no directed edge from the variable $X$, such as in the causal structure D or E, the importance will be zero. Thus, if an explanatory variable is independent of the target variable, the LEWIS explanatory score is zero because intervention on that explanatory variable cannot change the target variable. A similar argument is possible in the case of nonlinear causal structures to determine Nesuf.

\begin{table}[ht]
    \centering
    \caption{Mean of Nesuf in the case of linear causal structure}
    \label{tab:nesufL}
    \begin{tabular}{cccccc}
    \hline
          & A & B &C & D & E \\
         \hline
          $x$ &0.367
  &0.498
  &0.999
  &0  &0 \\
         $z$ &0.999  &0.999  &0.325
  &1  &1 \\
         \hline
    \end{tabular}
\end{table}
\begin{table}[ht]
    \centering
    \caption{Mean of Nesuf in the case of nonlinear causal structure}
    \label{tab:nesufN}
    \begin{tabular}{cccccc}
    \hline
          & A & B &C & D & E \\
         \hline $x$ &0.377
  &0.556
  &0.999
  &0  &0 \\
          $z$ &0.945  &0.999  &0.949
  &1  &1 \\
         \hline
    \end{tabular}
    
\end{table}
\subsection{Useful prior information on the causal structure}\label{priorinformation}
In Section~\ref{a1}, we observed that the score differed greatly depending on whether the explanatory variable could yield the predictive variable.
In particular, all explanatory scores will be 0 for the explanatory variables that are independent of the target variable. In addition, in case of a directed edge, or reverse causation, from the target variable to the explanatory variable, the target variable and explanatory variable are dependent. However, even if the value of the explanatory variables were changed, the value of the target variable would not be changed; thus, so the value of Nesuf becomes 0, and Suf and Nec provide no information. Therefore, we proposed the following prior information on the causal structure to compute the explanation scores:

\begin{enumerate}
\renewcommand{\labelenumi}{(\alph{enumi})}\label{katei}
    \item Target variable has the direct parent-child relationship with all explanatory variables, that is, there is a direct causal path from the explanatory variables to the target variable.
    \item Target variable is the sink variable, where a sink variable is a variable that does not cause any variable.
\end{enumerate}\par
\begin{figure}[t]
     \centering
     \includegraphics[scale=0.45]{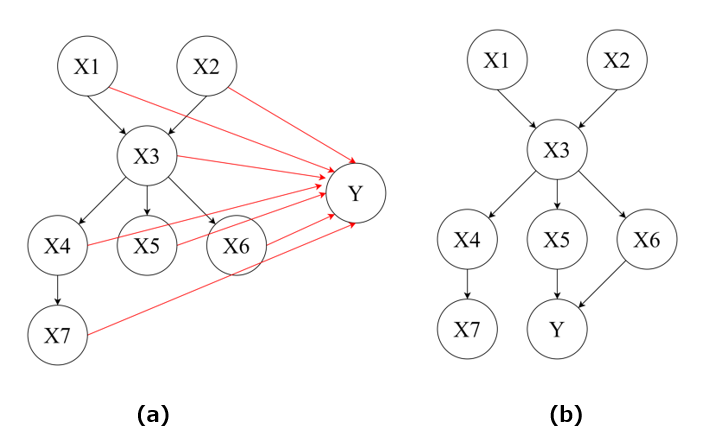}
     \caption{Prior information of causal structure. (a): Target variable Y has the direct parent-child relationship with all explanatory variables. (b): Target variable Y is the sink variable.}
     \label{fig:assum}
 \end{figure}

With prior information (a) (Figure~\ref{fig:assum}a), the target variable is dependent on all explanatory variables. In this prior information, the estimated score is adjusted by variables that satisfy the backdoor criteria; if they don't satisfy it, the intervention query is directly estimated, as in prior work. Hence if the explanatory variables and target variable are not independent, the estimated score will be at least the same as when no causal graph is used. This facilitates the performing of causal discovery using only explanatory variables.
In prior information (b) (Figure~\ref{fig:assum}b), we considered the influence of variables independent of the target variable and reverse causation, which could not be identified in prior information (a). In this prior information helps eliminate reverse causality from the target variable to the explanatory variables when performing causal discovery including the target variable. 

\section{Numerical experiment and evaluation}\label{S4}
This section describes the simulation and experimental results of explanations based on counterfactual probabilities through causal discovery using artificial data.
\subsection{Numerical experiment setting}
\begin{figure}[ht]
     \centering
     \includegraphics[scale=0.45]{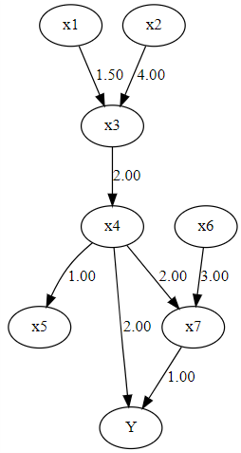}
     \caption{Causal graph in artificial data experiments
}
     \label{fig:exp2dag}
 \end{figure}

In the numerical experiment, we generated artificial data from the 8-variable causal graph shown in Figure~\ref{fig:exp2dag}, performed causal discovery based on prior information on the causal structure proposed in the previous section, and estimated the Nesuf for the predicted values. We performed evaluations considering the order of estimated the Nesuf and that of true Nesuf and the error of the estimated Nesuf value. We compared these results with the value of Nesuf estimated assuming that $P(Y|\mathit{do}(X))=P(Y|X)$, similar to that in case of \cite{LEWIS}. The artificial data to be generated were continuous and mixed data. Continuous variable values were generated from linear or non-linear function systems. For mixed data, only a linear system was used, the probability distribution of the error variables $X_1$ and $X_7$ was a Bernoulli distribution, and the probability distribution of the other error variables were a uniform or Gaussian distribution.
The model, sample size, discretization method, and the number of discretization bins are presented in Table~\ref{tab:exp1}.

 For the evaluation, we calculated the mean absolute error (MAE) with respect to the true Nesuf estimate. We used $\overline{\mathrm{MAE}}$, which is the average of the MAE
 \begin{equation}
     \mathrm{MAE} = \frac{1}{N}\sum_{i=1}^N |\mathrm{maxNesuf}_{X_i,\mathrm{true}} – \mathrm{maxNesuf}_{X_i,\mathrm{est}}|
 \end{equation}
for seven variables, $X_1,\,\ldots,\,X_7$, where $N$ is the number of experimental trials, $\mathrm{maxNesuf}_{X_i,\mathrm{true}}$ is the true value of $\mathrm{maxNesuf}(X_i)$, and $\mathrm{maxNesuf}_{x_i,\mathrm{est}}$ is the estimated value of $\mathrm{maxNesuf}(X_i)$.
The standard error of $\mathrm{maxNesuf}(X_i)$ was also evaluated.

 Here, we assume that $\mathbf{a}=\{a_1,\ldots,a_n\}$ and $\mathbf{b}=\{b_1,\ldots,b_n\}$ convert the value of the true Nesuf and the estimated Nesuf into a rank for $n$ variables, respectively. Defining $\mathbf{d}=\{d_1,\ldots,d_n\}$ by $d_i=a_i-b_i$, we can calculate Spearman's rank correlation coefficient \cite{sp} with
 \begin{equation}
     \mathrm{SPR} = 1-\frac{6\sum_{i=1}^nd_i^2}{n(n^2-1)},
     \label{sp}
 \end{equation}
where $\mathrm{SPR}$ attains the value $-1\leqq \mathrm{SPR} \leqq 1$. The closer it is to 1, the more the order of the estimated Nesuf matches the true order of feature importance. Moreover, the closer it is to $-1$, the more the order relationship is reversed.
In this experiment, the evaluation was based on the average MAE, the standard error and the average $\overline{\mathrm{SPR}}$ of 100 trials.

Next, for causal discovery, we used six methods tailored to specific functional forms. We implemented the causal discovery algorithm in Python. We used the causal-learn \cite{causal-learn} method for PC,  linear non-Gaussian acyclic model (LiNGAM) package \cite{package} for DirectLiNGAM, RESIT, and linear mixed (LiM) and causalnex\cite{causalnex} method for NOTEARS and NOTEARS-MLP.
 \begin{itemize}
     \item\textbf{PC} \cite{pc}: The PC algorithm is a causal discovery algorithm based on conditional independence. The estimation algorithm first removes undirected edges by testing conditional independence. The direction of the causal relationship is determined by using v-structures \cite{pearl2016} and orientation rule \cite{orien} for the remaining undirected skeleton. When using a PC to orient undirected edges, the directions may not be determined in the end, and the graph may be estimated as a partially oriented graph. In that case, our evaluation was based on the causal graph with all directed edges derived from the partially oriented graph that had the maximum (PC\_Max) or minimum (PC\_Min) Spearman's rank correlation coefficient. We used the Fisher-$z$ test to investigate the conditional independence.  
     \item \textbf{DirectLiNGAM} \cite{D-LiNGAM}: DirectLiNGAM is a causal discovery algorithm that expresses causal relationships using a linear structural equation model called LiNGAM \cite{LiNGAM}, which assumes that the graph is acyclic and that the probability distribution of the error terms is non-Gaussian. DirectLiNGAM estimates the causal structure by repeated regression and evaluating the independence of residuals of each variable.
     \item \textbf{RESIT} \cite{RESIT}: RESIT is a causal discovery algorithm used when the causal structure is nonlinear and the error variables are additive (Additive Noise Model: ANM). Similar to DirectLiNGAM, the algorithm estimates the causal direction by evaluating the independence of each variable and the residual of nonlinear regressions.
     \item \textbf{LiM} \cite{LiM}: LiM causal discovery algorithm extends LiNGAM to handle the mixed data that comprises both continuous and discrete variables. The estimation is performed by first globally optimizing the log-likelihood function on the joint distribution of data with the acyclicity constraint and then applying a local combinatorial search to output a causal graph.
     \item \textbf{NOTEARS} \cite{NOTEARS}: NOTEARS reformulates DAG structure learning as a continuous optimization problem over real matrices, avoiding combinatorial acyclicity constraints. It introduces a smooth characterization of acyclicity as an equality constraint $h(W)=0$ on the weighted adjacency matrix W. This equality-constrained program is solved using augmented Lagrangian methods and numerical optimizers. This method assumes that the probability distribution of the error term has equal variance. Empirically, it outperforms state-of-the-art methods, especially for linear.
     \item \textbf{NOTEARS-MLP}\cite{NOTEARS-MLP}: NOTEARS-MLP is that DAG structure learning algorithm extends NOTEARS to handle non-linear functional relationships using MLP which consists of hidden layer units and an activation that sigmoid function. Especially when the causal structure is an additive noise model, this method is identifiable that assuming the nonlinear function are three times differentiable and not linear in any of its arguments.
 \end{itemize}
 \par
 The following estimates were based on either prior information (a), prior information (b), or no prior information, denoted by (a), (b), and (0), respectively. PC and DirectLiNGAM can incorporate prior information (a) and (b). However, RESIT, LiM, NOTEARS and NOTEARS-MLP cannot incorporate prior information (b). In addition, we evaluated cases wherein causal discovery was performed without a causal graph (No graph). Table~\ref{cd} presents the experimental settings used in the numerical experiment.

 \begin{table}[ht]
     \centering
     \caption{Experimental settings for artificial data}
     \label{cd}
     
     \scalebox{0.9}{
     \begin{tabular}{|c|c|}
                \hline 
                Data & \{continuous, mix\} \\
                Function&\{linear, nonlinear\}\\
                Probability of error variables &\{uniform, Gaussian\} \\
                 Sample size &5000\\
                Discretization method& equal-width discretization\\
                Number of discretization bins&10\\
                Model & catboost\\
                Prior knowledge& \{(0), (a), (b), (No graph)\}\\
                \hline         
            \end{tabular}
            }        
 \end{table}

\subsection{Results of experiment}
The experimental results of $\overline{\mathrm{MAE}}$ $\pm$ standard error and $\overline{\mathrm{SPR}}$ are presented in Tables~\ref{tab:LU}–\ref{tab:LGM}. First, regarding the linear causal structure in Tables~\ref{tab:LU} and \ref{tab:LG}, when the causal discovery matched the true graph, the $\overline{\mathrm{MAE}}$ approached 0 and the $\overline{\mathrm{SPR}}$ approached 1. DirectLiNGAM assumed that the functional relationship was linear and that the probability distribution of the error terms was non-Gaussian. In such cases, $\overline{\mathrm{MAE}}$ was smaller and the $\overline{\mathrm{SPR}}$ was greater than that in the case No graph (Table~\ref{tab:LU}). As shown in Table~\ref{tab:LG}, when a Gaussian distribution was assumed with no prior information or No graph, the $\overline{\mathrm{MAE}}$ was large and the $\overline{\mathrm{SPR}}$ was small for DirectLiNGAM. However, these scores were improved with prior information (a) and (b). Furthermore, the PC and NOTEARS algorithm consistently achieved high accuracy for both distributions.
In all cases with a linear structure, we could estimate the true value and order of Nesuf scores better than that case when No graph was assumed.
\begin{table}[htbp]
    \centering
    \caption{Results of linear and uniform distribution}
    \label{tab:LU}
    {
    \begin{tabular}{ccc}
    \hline
         Method and prior information&$\overline{\mathrm{MAE}}$&$\overline{\mathrm{SPR}}$  \\
         \hline DirectLiNGAM (0)& \bf{0.0032 ± 0.0197
}&\bf{0.9889}
\\
DirectLiNGAM (a)& 0.1700 ± 0.0023
 &0.6140\\
          DirectLiNGAM (b)& \bf{0.0032 ± 0.0197
}&\bf{0.9889}
\\
          
          PC\_Max (0) &0.0633 ± 0.0595 &0.9007\\
          PC\_Min (0)&0.0953 ± 0.1127 &0.8507\\
          PC\_Max (a)&0.1709 ± 0.0030 &0.6117\\
          PC\_Min (a)&0.1709 ± 0.0030 &0.6117\\
          PC\_Max (b)&0.0092 ± 0.0249 &0.9817\\
          PC\_Min
         (b)&0.0703 ± 0.0656 &0.9082\\
          NOTEARS (0)&0.1708 ± 0.0030  &   0.6117 \\
          NOTEARS (a)&0.1708 ± 0.0030  &   0.6117 \\
           No graph &0.1709 ± 0.0030 &0.6117\\
         \hline 
    \end{tabular}
    \par
    \small Values presented in bold indicate the best results.
    }
    
\end{table}
\begin{table}[htbp]
    \centering
    \caption{Results of linear and Gaussian distribution}
    \label{tab:LG}
    {
    \begin{tabular}{ccc}
             \hline
             Method and prior information&$\overline{\mathrm{MAE}}$&$\overline{\mathrm{SPR}}$  \\
             \hline DirectLiNGAM (0)&0.3605 ± 0.1290
&0.3307
\\
DirectLiNGAM (a)&0.1910 ± 0.0100
&0.5920
\\
              DirectLiNGAM (b)&0.2846 ± 0.1080
 &0.7403
\\
              
              PC\_Max (0)&0.0981 ± 0.0461 &0.8796\\
              PC\_Min (0)&0.1186 ± 0.0920 &0.8492\\
              PC\_Max (a)&0.1912 ± 0.0097 &0.5550\\
              PC\_Min (a)&0.1912 ± 0.0097 &0.5550 \\
              PC\_Max (b)&
              \bf{0.0088 ± 0.0287 }&\bf{0.9764}\\
              PC\_Min (b)&0.1013 ± 0.0584 &0.8971\\
              NOTEARS (0)&0.4162 ± 0.0804       &0.6739\\
          NOTEARS (a)&0.1913 ± 0.0095
          &0.5550 \\
               No graph&0.1912 ± 0.0097 &0.5550\\
             \hline
        \end{tabular}
        }
    
\end{table}

Tables~\ref{tab:NU} and~\ref{tab:NG} present the results in the case of a nonlinear causal structure. When the error term was uniform distribution,  RESIT with no prior information and PC with prior information yielded smaller $\overline{\mathrm{MAE}}$ values and greater $\overline{\mathrm{SPR}}$  values than that for No graph in Table~\ref{tab:NU}.
 For the Gaussian distribution causal structure as shown in Table~\ref{tab:NG}, NOTEARS-MLP with prior information had the highest performance. However, in this experimental setup, the performance of all methods was lower compared to other setups. The reason may be that monotonicity, another assumption of LEWIS, is not satisfied in nonlinear cases.
\begin{table}[htbp]
    \centering
    \caption{Results of nonlinear and uniform distribution}
        \label{tab:NU}
    {
    \begin{tabular}{ccc}
        \hline
             Method and prior information&$\overline{\mathrm{MAE}}$&$\overline{\mathrm{SPR}}$ \\
             \hline RESIT (0) &0.0250 ± 0.0190
&\bf{0.8114}
\\
              RESIT (a)&0.0334 ± 0.0059
 &0.7285
\\
              PC\_Max (0)&0.0398 ± 0.0273 &0.5540\\
              PC\_Min (0)&0.0398 ± 0.0273 &0.5540\\
              PC\_Max (a)&0.0337 ± 0.0064 &0.7781\\
             PC\_Min (a)&0.0337 ± 0.0064 &0.7781\\
              PC\_Max (b)&\bf{0.0183 ± 0.0226} &0.7781\\
             PC\_Min (b)&\bf{0.0183 ± 0.0226} &0.7781\\
              NOTEARS-MLP (0)
              &0.0521 ± 0.032     &   0.6053\\
              NOTEARS-MLP (a)
              &0.0337 ± 0.0063       &0.7067\\
              No graph&0.0337 ± 0.0064
&0.7067
\\
             \hline  
        \end{tabular}
        }
\end{table}
\begin{table}[htbp]
    \centering
    \caption{Results of nonlinear and Gaussian distribution}
            \label{tab:NG}
    {
    \begin{tabular}{ccc}
              \hline
             Method and prior information&$\overline{\mathrm{MAE}}$&$\overline{\mathrm{SPR}}$  \\
             \hline RESIT (0)&\bf{0.0292 ± 0.0115}
 &0.3296
\\
              RESIT (a)&0.0377 ± 0.0128
 &0.3235
\\
              PC\_Max (0)&0.0323 ± 0.0177 &0.1853\\
              PC\_Min (0)&0.0325 ± 0.0179 &0.1810\\
              PC\_Max (a)&0.0372 ± 0.0128
              &0.3317\\            
             PC\_Min (a)&0.0372 ± 0.0128 &0.3317\\
             PC\_Max (b)&0.0325 ± 0.0179 &0.1810\\
             PC\_Min (b)&0.0325 ± 0.0179 &0.1810\\
             NOTEARS-MLP (0)
              &0.0575 ± 0.0098       &0.3364 
              \\
              NOTEARS-MLP (a)
              &0.0382 ± 0.0130       
              &\bf{0.3857}\\
                No graph&0.0372 ± 0.0128 &0.3317\\
             \hline
        \end{tabular}
        }           
\end{table}

Finally, Tables~\ref{tab:LUM} and \ref{tab:LGM} present the results of the mixed data. In Table~\ref{tab:LUM}, the assumptions of LiM were satisfied, thus, the $\overline{\mathrm{MAE}}$ was smaller and the $\overline{\mathrm{SPR}}$ was larger than that when no causal discovery was performed. In addition, in Table~\ref{tab:LGM}, the LiM assumption that the probability distribution of the error terms was non-Gaussian was not satisfied, the score was the same as the case of No graph. Even in the case of linear mixed data, it can be said that the true Nesuf can be partially restored using the proposed prior information of causal structure.

\begin{table}[ht]
    \centering
    \caption{Results of linear and uniform distribution in mix data}
        \label{tab:LUM}
    \begin{tabular}{ccc}
        \hline
             Method and prior information&$\overline{\mathrm{MAE}}$&$\overline{\mathrm{SPR}}$  \\
             \hline LiM (0)&\bf{0.1234 ± 0.0287}&0.6571\\
              LiM (a)&0.1380 ± 0.0195 &\bf{0.6892}\\
              No graph &0.1741 ± 0.0145&0.4292\\
             \hline  
        \end{tabular}
\end{table}
\begin{table}[ht]
    \centering
    \caption{Results of linear and Gaussian distribution in mix data}
    \label{tab:LGM}
    \begin{tabular}{ccc}
              \hline
             Method and prior information&$\overline{\mathrm{MAE}}$&$\overline{\mathrm{SPR}}$  \\
             \hline LiM (0)&0.1726 ± 0.0184&0.4350\\
              LiM (a)&\bf{0.1540 ± 0.0234}&\bf{0.4635}\\
                No graph&0.1546 ± 0.0028&0.4603\\
             \hline
        \end{tabular}
    
\end{table}

This result confirms that prior information (a) is at least the same and more than that as when no causal graph is used. Futhermore prior information (0) and (b) provide relatively high performance compared to the case with No graph. On the other hand, depending on the nature of the data and the causal discovery method, it may be worse than the case with No graph, so it is necessary to consider multiple causal discovery methods.

\section{Application to real data}\label{S5}
In this section, we demonstrate the effectiveness of this framework for the case where the causal graph is unknown by applying our method to real data.
\subsection{Dataset and preprocessing}
We applied our method to the anonymized credit rating data of $14\,018$ business customers provided by the Shiga Bank, Ltd.
A credit rating was assigned by a bank to a debtor based on the analysis of its financial statements \cite{saimu}.
Although there were several grades in the credit rating, we simplified the grades to high (excellent) and low (poor), which facilitated its modeling as a binary classification problem.
We used the industry type, amount of capital stock, number of employees, most recent annual sales, and total liabilities and equity as the explanatory variables.
We performed equal-frequency discretization with 10 bins because the capital stock, number of employees, most recent annual sales, and total liabilities and equity had highly skewed distributions. These discretized variables are ordinal measures with enough many levels and can be seen as continuous variables. As a machine learning model, we used Random Forest, which is an algorithm that performs classification by combining the results of multiple decision trees constructed from randomly selected learning data and explanatory variables; it is an ensemble learning algorithm \cite{Ra}. Random Forest was expected to have sufficient predictive accuracy for real data in this study.

Although the data involved mixed data where only the industry type is discrete variable and the others are regarded as continuous variables, the industry type could not be caused by the other explanatory variables. Thus, we used DirectLiNGAM for continuous variables, assuming that industry type was an exogenous discrete variable that affected all other variables. In this case, DirectLiNGAM can handle discrete variable similarity to the conventional structural equation modeling \cite{B-sem}. Because we aimed at a quantitative analysis of all explanatory variables, we conducted our analysis assuming that the target variable exhibited a direct parent-child relationship with all explanatory variables in the causal structure, that is, prior information (b).

\subsection{Analysis results and discussion}
\begin{figure}[htbp]
    \centering
    \includegraphics[scale=0.6]{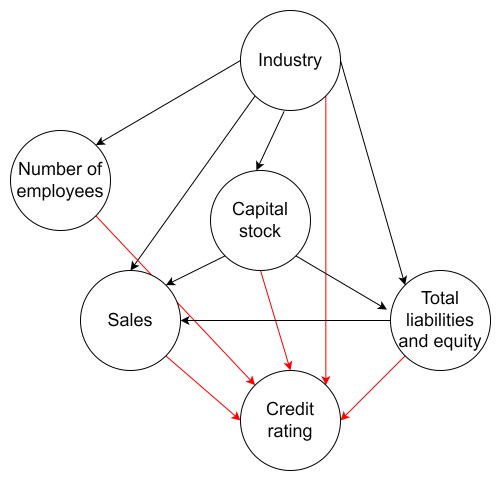}
    \caption{Causal graph estimated by DirectLiNGAM 
 (black lines). Prior information on the causal structure (red lines).}
    \label{fig:causal_graph}
\end{figure}

The black lines of Figure~\ref{fig:causal_graph} show the results with DirectLiNGAM for the entire dataset. The interpretation of the causal direction of the results is as follows. The causal direction from capital stock to total liabilities and equity was consistent with domain knowledge because the total liabilities and equity are the sum of debt and equity on the balance sheet.
In addition, the causal direction from capital stock to sales was consistent with the domain knowledge that companies conducted business activities based on capital and that this influenced sales. These variables can affect the credit rating on the basis of prior information of the causal structure in Figure~\ref{fig:causal_graph} (red lines).

\par

\begin{figure}[htbp]
    \centering
    \includegraphics[scale=0.6]{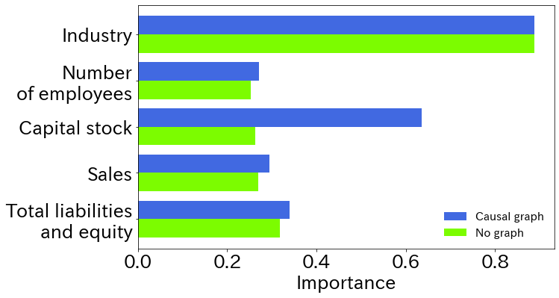}
    \caption{Nesuf estimated from causal graph and No graph}
    \label{fig:fp}
\end{figure}

Next, Nesuf estimated using the causal graph and No graph is shown in Figure~\ref{fig:fp}. The importance ranking of each of these variables obtained by LEWIS can be explained as follows using the causal graph shown in Figure~\ref{fig:causal_graph}.
The industry type was an exogenous variable that affected the other four variables on the causal graph. If the industry type were to be changed, the values in the other four variables would also change, which is likely to affect the final predicted value. Thus, its importance in LEWIS is the greatest. Similarly, the value of capital stock changed the value of the sales and total liabilities and equity, and the total liabilities and equity changed the value of sales. The number of employees and sales were less important because changing these values did not change the values of other variables. However, the Nesuf scores estimated with No graph were smaller, and those of the capital stock in particular were significantly smaller, which contradicted domain knowledge.

\begin{figure}[ht]
    \centering
    \includegraphics[scale=0.53]{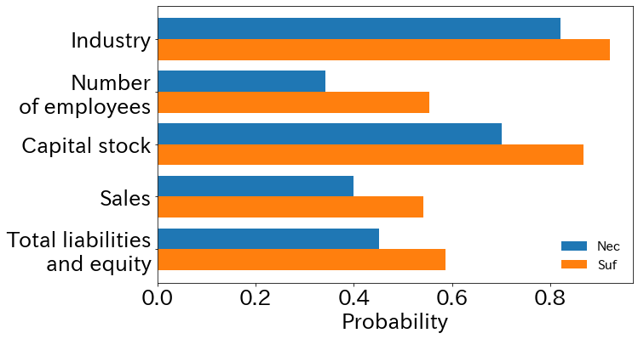}
    \caption{Reversal probability score estimated from the causal graph. Nec (blue) is the probability that the prediction would change by lowering the value of that variable for a company whose rating is predicted to be high. Suf (orange) is the probability that the prediction would change by increasing the value of the variable for a company whose rating is predicted to be low.}
    \label{fig:反転}
     
\end{figure}

The LEWIS reversal probability score is shown in Figure~\ref{fig:反転}. Overall, the Suf score (orange) was high, and there was a high probability that changing the value of the variable would change the company from a low-rated one to a high-rated one. Industry type also significantly affected the predicted value if it were changed. Reversal probabilities can assist the decision-making about which variables are likely to change the rating for a low-rated company and can quantify the company's strengths and weaknesses based on each score.

\section{Conclusion}\label{S6}
This study proposed a new causal XAI framework that combined causal structure information and causal discovery without the knowledge of the causal graph. We analyzed the global explanation scores by using counterfactual explanations based on the causal structure and proposed prior information on the causal structure. Numerical experiments demonstrated the possibility of estimating the global explanatory score and the order of the true feature importance even if the causal graph was not fully known. By applying our method to real data, we demonstrated the usefulness of the proposed framework even if the causal graph is unknown. 

As an extension of LEWIS, a method for multi-class classification was proposed \cite{LEWIS}. However, whether proposed prior information on causal structure is valid even for the multi-class classification remains an open question. 
\section*{Acknowledgment}
We would like to express our gratitude to Shiga Bank, Ltd. for their valuable comments on this study and for providing us with the data used. This work was partially supported by JSPS KAKENHI 20K11708 and JST CREST JPMJCR22D2.

\end{document}